\pdfoutput=1

\documentclass[11pt]{article}

\usepackage[]{acl}

\usepackage{times}
\usepackage{latexsym}
\usepackage{multicol}
\usepackage{multirow}

\usepackage[T1]{fontenc}

\usepackage[utf8]{inputenc}

\usepackage{microtype}

\usepackage{inconsolata}

\usepackage{graphicx}

\title{Enhancing Investment Opinion Ranking through Argument-Based Sentiment Analysis}

\author{Chung-Chi Chen,\textsuperscript{\rm 1}
    Hen-Hsen Huang,\textsuperscript{\rm 2}
    Hsin-Hsi Chen \textsuperscript{\rm 3}  \\
    \textbf{Hiroya Takamura,\textsuperscript{1} Ichiro Kobayashi,\textsuperscript{4}  Yusuke Miyao\textsuperscript{5}}
    \\
    \textsuperscript{\rm 1} AIST, Japan \textsuperscript{\rm 2}    Institute of Information Science, Academia Sinica, Taiwan \\
    \textsuperscript{\rm 3} Department of Computer Science and Information Engineering, \\ 
   National Taiwan University, Taiwan \textsuperscript{4} Ochanomizu University, Japan \textsuperscript{5} University of Tokyo, Japan \\
    c.c.chen@acm.org,
    hhhuang@iis.sinica.edu.tw,
    hhchen@ntu.edu.tw, \\takamura.hiroya@aist.go.jp,koba@is.ocha.ac.jp, yusuke@is.s.u-tokyo.ac.jp}

\begin{document}
\maketitle
\begin{abstract}
In the era of rapid Internet and social media platform development, individuals readily share their viewpoints online. The overwhelming quantity of these posts renders comprehensive analysis impractical. This necessitates an efficient recommendation system to filter and present significant, relevant opinions. Our research introduces a dual-pronged argument mining technique to improve recommendation system effectiveness, considering both professional and amateur investor perspectives. Our first strategy involves using the discrepancy between target and closing prices as an opinion indicator. The second strategy applies argument mining principles to score investors' opinions, subsequently ranking them by these scores. Experimental results confirm the effectiveness of our approach, demonstrating its ability to identify opinions with higher profit potential. Beyond profitability, our research extends to risk analysis, examining the relationship between recommended opinions and investor behaviors. This offers a holistic view of potential outcomes following the adoption of these recommended opinions.

\end{abstract}

\section{Introduction}

In recent years, online platforms catering to diverse interests have proliferated, ranging from personal updates on Twitter and Instagram to professional discourse on MathOverflow and Stack Overflow. Specialized platforms also exist for sharing perspectives on specific topics, such as iDebate\footnote{\url{https://idebate.org/}} for user debates and various e-commerce sites for product reviews. Among these, platforms focusing on future-oriented user predictions, especially in the investment domain, play a unique role. Sites like Yahoo Finance\footnote{\url{https://finance.yahoo.com/research/}} and Reddit's WallStreetBets\footnote{\url{https://www.reddit.com/r/wallstreetbets/}} are popular for sharing professional and amateur investment opinions. However, as \citeauthor{xiao2015does}~\cite{xiao2015does} notes, a mere 20\% of investors see profits from stock market activities, underscoring the challenge of discerning valuable investment opinions amidst a deluge of information. This situation calls for the development of an effective recommendation system to assist investors in identifying advantageous opinions.

Previous research has primarily focused on predicting stock price movements using publicly available opinions~\cite{sawhney-etal-2020-deep,xu-cohen-2018-stock,bollen2011twitter}. However, few have concentrated on extracting the subset of opinions that are likely to be profitable. Current methods rely on analyzing either the characteristics of the authors or the content of their posts. Author-centric analysis, despite its usefulness in identifying potential sources of profitable insights, faces several challenges, such as the lack of historical data for new users, the anonymity and account variability on social media, and the inconsistency of user information across platforms.
To address these challenges, this paper proposes a text-centric approach to analyze opinions and identify those likely to result in profitable investments. This involves using historical data to train models for predicting profitable outcomes and extracting linguistic features to sort out useful opinions. We introduce a novel method that leverages the concept of argument mining to analyze opinions. This approach decomposes opinions into premises and claims, examining their interrelations to provide a more nuanced understanding of the opinion expressed.

Argument mining, as described by Mochales and Moens~\cite{mochales2011argumentation}, is the process of extracting and identifying the argumentative structure of a given text, particularly in subjective and persuasive narratives. Investment opinions, which forecast market futures, inherently contain subjective viewpoints of investors. To persuade readers, it is essential to support claims with evidence (premises). This paper goes beyond merely distinguishing between premises and claims in a narrative; we introduce the concept of measuring the strength of an argument. 
A novel aspect of our methodology is the use of price targets set by professional analysts as proxies for argument strength. These targets, representing the analysts' future price estimations, enable the quantification of the argument's strength.  Additionally, this study investigates the influence of objective evidence and subjective opinion in forming investment recommendations. Through argument mining, we distinguish between factual statements and personal views within an opinion, thereby enhancing the analysis. Our approach, termed argument-based opinion analysis, presents a novel perspective in the domain of investment opinion recommendation.

Our research also extends into the realm of professional behavior, examining changes in analysts' views and trading patterns among professional traders. We find that opinions ranked highly by our method are more correlated to analysts' revisions and traders' decisions. This aspect of professional behavior, though crucial, has been largely overlooked in previous studies.

In summary, the contributions of this work are threefold:
\begin{enumerate}
    \item Introduction of a novel approach for assessing strength degree of opinion using price targets set by professional analysts.

    \item Development and evaluation of various strategies based on argument mining notions for investment opinion recommendation.

    \item Comprehensive analysis of the implications of differently ranked opinions on profitability, risk, and professional trading behaviors.

\end{enumerate}

\section{Related Work}
Investment decision-making based on online opinions has been a topic of long-standing discussion. The majority of studies have harnessed the 'wisdom of the crowd' by aggregating available opinions~\cite{sawhney-etal-2020-deep,xu-cohen-2018-stock,bollen2011twitter}, with few exploring the utility and relevance of individual opinions. In this paper, we align with the latter's perspective, proposing a novel direction for the assessment of individual opinions. 

Historically, research has examined this issue from two angles: author-based and content-based. Bar-Haim et al.~\cite{bar-haim-etal-2011-identifying} focused on identifying experts based on their past performance. Tu et al.~\cite{tu2018investment} prioritized authors' social popularity to sift through investors' opinions. However, these methods encounter a significant 'cold start' problem where new users lack historical data and users can manipulate their online presence through account changes or deletion of past posts. Moreover, it can be challenging to obtain the necessary author-specific data in various scenarios. For instance, the social popularity of professional analysts is not readily available for assessing their opinions. As a result, we propose a content-centric approach that can be universally applicable.

Related studies have also examined opinions based on content. Zong et al.~\cite{zong-etal-2020-measuring} used BERT~\cite{devlin-etal-2019-bert} to classify the accuracy of analyst reports according to company earnings forecasts and actual earnings. This was treated as a binary classification task, wherein a given report was either in the top or bottom $K$ groups, selected from 16,044 reports. However, we contend that such a coarse setting is unsuitable for recommending a manageable number of reports for investors. Furthermore, social media platforms typically lack historical performance records, making it challenging to compile sizeable training data. In contrast, our method can rank opinions on a continuous scale (from 0 to 1) using a limited number of reports with automatically annotated information.

Chen et al.~\cite{chen2021evaluating} put forth an approach to identify high-predictive-power opinions on social media platforms, wherein the predictive power of posts was determined by the similarity of their sentences to those written by professional analysts. However, this approach faces a significant issue in that it cannot rank professional investors' opinions, as all sentences in professional reports are authored by professionals. We offer a novel method to overcome this limitation. Our approach effectively ranks both professional and amateur opinions, and our experimental results demonstrate its superior performance in identifying opinions leading to higher potential profit.

\section{Task Formulation and Dataset}
\subsection{Investor Opinion Recommendation}
This paper is tasked with discerning the potentially profitable investment opinions from a vast array of perspectives. Rather than merely sifting through the plethora of opinions, our objective is to pinpoint and recommend those with the highest potential for yielding profitable outcomes to investors. Considering that most investor opinions only provide information on initiating a trading position, such as whether to go long or short on a stock, but don't suggest when to close the position, assessing the actual return achievable by following the opinion on the market becomes challenging. As such, we adopt the approach of previous work~\cite{chen2021evaluating}, using the Maximum Possible Profit (MPP) as the evaluation metric for potential return. This approach circumvents the problem of determining when to close the position and provides a reasonable estimation of the potential return from an opinion. Below are the equations used to calculate the MPP of both bullish and bearish opinions for a stock on day $t$:

\begin{equation}\label{eq:mpp_bullish}
    MPP_{\mathit{bullish}} = \max_{t+1 \le i \le t+T} \frac{H_i - O_{t+1}}{O_{t+1}}
\end{equation}

\begin{equation}\label{eq:mpp_bearish}
    MPP_{\mathit{bearish}} = \min_{t+1 \le i \le t+T}\frac{O_{t+1} - L_i}{O_{t+1}}
\end{equation}

In the above equations, $O_{t}$ represents the opening price on day $t$, with $H_t$ and $L_t$ denoting the highest and lowest prices on day $t$ respectively. Following \citeauthor{chen2021evaluating}~\cite{chen2021evaluating}, we set $T$ as 60 trading days, or roughly three months. Ultimately, our goal is to devise a method to identify opinions that could lead to a higher MPP.

\subsection{Set of Investors' Opinions}
\label{sec:Set of investors' opinions}
Contrary to prior studies that focused solely on either formal reports~\cite{zong-etal-2020-measuring} or social media posts~\cite{chen2021evaluating}, our work proposes a method capable of handling both. Consequently, our experiments involve two sets of opinions: professional reports and social media posts. This section details these opinion sets. It should be noted that these sets are used exclusively for evaluating recommendation approaches and not during any training steps.

We assembled a collection of 2,280 professional analysis reports from the Bloomberg Terminal. Written in Chinese, these reports delve into stocks listed on the Taiwan stock market, covering a total of 513 different stocks. The release date for each report, indicated in the file name, enables us to easily ascertain $t$. Once $t$ is known, we align it with the price data based on date information, allowing us to calculate the MPP for each report. This collection is henceforth referred to as the PAR set.

A set of social media posts was obtained from \citeauthor{chen2021evaluating}~\cite{chen2021evaluating}. Written in Chinese, these posts were sourced from PTT, a well-known social media platform in Taiwan.\footnote{\url{https://www.ptt.cc/bbs/Stock/index.html}} \citeauthor{chen2021evaluating}~\cite{chen2021evaluating} manually annotated the MPP of each post, considering the sentiment (bullish/bearish) expressed in these posts and the stocks discussed. This collection is referred to as the SMP set moving forward. Our experiment aims to recommend opinions from the PAR and SMP sets that could lead to a higher MPP.

\section{Our Approach}
\label{sec:Approach}
Our proposed method utilizes a three-step approach which incorporates two concepts to evaluate the given opinion: strength degree estimation and argument-based analysis. 
Initially, we estimate the strength degree of each sentence within the given opinion. Subsequently, we categorize each sentence as claim, premise, or others. Following this, we identify the relationships between claims and premises, thereby discerning which premise supports a given claim and determining the strength degree of the arguments. Finally, we recommend opinions based on the extracted strength degree and argument features. Detailed explanations of our proposed method are provided in this section.

\subsection{Strength Degree}
\label{sec:sentiment degree}

To acquire the strength degree ($SD$), we suggest using the difference between the price target ($PT$) and the most recent close price as a proxy. Equation~\ref{eq:sentiment} formulates our method for obtaining $SD$.

\begin{equation}
\label{eq:sentiment}
    SD = \frac{PT - C_{t^r}}{C_{t^r}}
\end{equation}

Here, $C_{t^r}$ is the close price at the time the report was released on date $t^r$. 

For generating a dataset with fine-grained strength degrees, we collate an additional 628 professional analysis reports from Bloomberg Terminal. These reports do not overlap with the ones introduced in Section~\ref{sec:Set of investors' opinions}. We extract the $PT$, $C_{t^r}$, and the content from these reports. As a result, we obtain 37,147 sentences from these reports, with sentences from the same report being labeled with the same $SD$. 

These sentences are segregated into two groups based on whether the $SD$ of the sentence is higher than the average $SD$ (22.51\%) of all sentences or not. We then fine-tune BERT-Chinese~\cite{devlin-etal-2019-bert} to perform this binary classification task. Using 80\% of the instances for training and the remaining for testing, we develop a model that achieves micro- and macro-averaged F-scores of 66.78\% and 62.10\%, respectively. 
This fine-tuned model is then employed to predict each sentence in the PAR and SMP sets. We use the probability that the given sentence will have $SD$ higher than the average as the fuzzy strength degree ($FSD$), which ranges from 0 to 1. Hence, all sentences in PAR and SMP sets are assigned an $FSD$.

\begin{table}[t]
  \centering
      \resizebox{\columnwidth}{!}{
    \begin{tabular}{cl}
    \hline
    $FSD$    & \multicolumn{1}{c}{Sentence} \\
    \hline
    0.93  & Expected to grow in both old and new businesses \\
    0.88  & Driven by memory and Indian factories \\
    0.85  & Profits in 2019 will show explosive growth \\
    \hline
    0.59  & the company implements epidemic prevention measures \\
    0.50  & companies are expected to introduce new products \\
    \hline
    0.15  & because of Chinese manufacturers bidding for orders \\
    0.04  & Considering the slow recovery speed of operations \\
    0.01  & Operation still hasn't got rid of the downturn \\
    \hline
    \end{tabular}%
    }
    \caption{Sentences and the corresponding $FSD$.}
  \label{tab:fsd and the corresponding sentence}%
\end{table}%

Table~\ref{tab:fsd and the corresponding sentence} displays some instances and the corresponding $FSD$. It is observed that the higher $FSD$ is stronger to support the bullish stance. This observation suggests that the proposed approach can effectively estimate the strength degree.

\subsection{Argument-based Opinion Analysis}
Argument mining has been a major focus of research over the past decade~\cite{lawrence2020argument,cabrio2018five}. Fundamental tasks such as argument detection are crucial in analyzing narratives like debate discourse and persuasive essays~\cite{shnarch-etal-2020-unsupervised,levy-etal-2018-towards,shnarch-etal-2018-will,rinott-etal-2015-show,chen2020numclaim}. Furthermore, the linking of arguments, claims, and premises can form a structure that provides a better understanding of the author's strategy and narrative flow~\cite{li-etal-2020-exploring-role,beigman-klebanov-etal-2016-argumentation,wachsmuth-etal-2016-using,chen2021evaluating}. Building on these previous works, we incorporate both argument detection and argument relation linking tasks in our analysis of investors' opinions.

We utilize the pretrained BERT-Chinese models proposed by \citeauthor{chen2021evaluating}~\cite{chen2021evaluating} for argument detection and argument relation linking. These models yield macro-averaged F-scores of 79.86\%, 57.69\%, and 56.96\% for claim detection, premise detection, and relation linking, respectively. We use these models to categorize the sentences in the PAR and SMP datasets as either claims or premises. The pretrained relation linking model is then used to identify whether a sentence designated as a premise supports an identified claim. 

This approach allows us to dissect an opinion across multiple dimensions. For instance, consider the sentences ``Profits in 2019 will show explosive growth'' and ``because of Chinese manufacturers bidding for orders'' from Table~\ref{tab:fsd and the corresponding sentence}. While the former is a claim and the latter a premise, do they serve the same function in shaping investors' opinions? Given that the claim predicts future earnings or operations and the premise describes occurred or plausible events, we aim to determine which part should receive more focus in recommending investors' opinions. In addition, we aim to investigate whether sentences not related to the claim (like greetings) should be included or excluded from the analysis. 

In our experiments, we explore these research questions using the assigned argument labels and propose the following strategies to evaluate the strength degree of each opinion in the PAR and SMP datasets:

\begin{itemize}
\small
    \item \textit{AllSent}: Average the $FSD$ of all sentences in the opinion.
    
    \item \textit{AllArg}: Average the $FSD$ of the sentences labeled as either a claim or a premise.
    
    \item \textit{ClaimOnly}: Average the $FSD$ of the sentences labeled as a claim.

    \item \textit{PremiseOnly}: Average the $FSD$ of the sentences labeled as a premise.
    
    \item \textit{KeyPremise}:  Use the maximum $FSD$ among all premises linked to the claims.
\end{itemize}
In summary, we investigate various strategies to aggregate sentence-level $FSD$ to opinion-level $FSD$ and rank investors' opinions based on this opinion-level $FSD$.

\section{Experiment}
\subsection{Recommendation Results in PAR set}
In accordance with \citeauthor{zong-etal-2020-measuring}~\cite{zong-etal-2020-measuring}, BERT demonstrates superior performance in classifying accurate and inaccurate reports. Since such a classification method is not designed for ranking reports, we suggest several alternative approaches to utilize the model's predictions:
\textbf{BERT-Conf}: This method classifies reports into higher Mean Percentage Profit (MPP) (Top 50\%) and lower MPP (Bottom 50\%) groups using BERT. To sequence the reports, we use the calculated probability that a given report belongs to the top 50\% as a proxy, sorting the reports based on this probability.
\textbf{BERT-Reg}: This strategy employs BERT to predict the MPP of a given report, treating the task as a regression problem. The predicted MPP is used to rank the reports.
\textbf{Megzi-FinBERT-Reg}: This approach replaces BERT in BERT-Reg with Megzi-FinBERT~\cite{zhang2021mengzi}, which is pre-trained using financial news and reports.

For a fair comparison, we incorporate the additional 628 reports discussed in Section~\ref{sec:sentiment degree} in the training process. Additionally, the method proposed by \citeauthor{chen2021evaluating}~\cite{chen2021evaluating}, which is based on the principle of counting expert-like sentences in social media posts, is unsuitable for ranking professional reports because all sentences in the PAR set are expert-written.

\begin{table}[t]
  \centering
  \resizebox{\columnwidth}{!}{
    \begin{tabular}{lrrrr}
    \hline
    \multirow{2}[0]{*}{Strategy} & \multicolumn{2}{c}{Top} & \multicolumn{2}{c}{Last} \\
          & 10th Decile & 9th Decile & 2nd Decile & 1st Decile \\
    \hline
    BERT-Conf~\cite{zong-etal-2020-measuring} & 11.68\% & 12.42\% & 15.02\% & 15.14\% \\
    BERT-Reg~\cite{devlin-etal-2019-bert} & 12.96\% & 12.58\% & 13.23\% & 12.05\% \\
    Mengzi-FinBERT-Reg~\cite{zhang2021mengzi} & 10.97\% & 12.08\% & 14.96\% & 15.29\% \\
    \textit{AllSent}   & 15.25\% & 14.53\% & 12.75\% & 11.93\% \\
    \textit{AllArg}   & 14.36\% & \textbf{14.75\%} & 12.73\% & 11.93\% \\
    \textit{ClaimOnly}   & 14.27\% & 14.51\% & 12.78\% & 11.39\% \\
    \textit{PremiseOnly}   & 14.51\% & 14.35\% & 9.39\% & 2.46\% \\
   \textit{KeyPremise}   & \textbf{15.59\%} & 14.71\% & \textit{5.01\%} & \textit{1.46\%} \\
    \hline
    \end{tabular}%
    }
        \caption{Averaged MPP of the sorted professional reports under different strategies. For the top-ranked reports, we bold the highest one; and for the ranked-last reports, we use italic to mark the worst one. }
  \label{tab:Results of sorting professional reports}%
\end{table}%

We adopt the evaluation method in \citeauthor{chen2021evaluating}~\cite{chen2021evaluating} to report the average MPP in the 10th and 9th deciles. We also report the last two deciles to demonstrate whether the proposed method can highlight opinions with higher MPP and filter out those with lower MPP. Table~\ref{tab:Results of sorting professional reports} displays the average MPP of the top-ranked and lowest-ranked reports based on different strategies. The p-value of the Friedman test is 0.009, suggesting a significant difference among the ranking results of these strategies with a 95\% confidence level.

Several findings can be drawn from Table~\ref{tab:Results of sorting professional reports}:
First, the baseline model, BERT-Conf~\cite{zong-etal-2020-measuring}, appears ineffective for ranking reports as the reports with a higher probability in the top 50\% group yield a lower MPP. For a more detailed analysis, we calculate the average MPP of all opinions categorized as the top 50\% and bottom 50\% groups, comparing the results with the best-performing strategy, \textit{KeyPremise}.
Second, we observe a correlation between the proposed $FSD$ and potential profit (MPP). Regardless of the strategies adopted, the top-ranked opinions consistently lead to higher potential profit than the lowest-ranked opinions. This implies that the proposed $FSD$ can be effectively used to rank investors' opinions.
Third, the ordering results of \textit{AllSent}, \textit{AllArg}, and \textit{ClaimOnly} are strikingly similar in both top-ranked and lowest-ranked opinions. Both \textit{PremiseOnly} and \textit{KeyPremise} effectively filter out the lowest-ranked opinions. The only difference between \textit{PremiseOnly} and \textit{AllSent} is that \textit{PremiseOnly} exclusively uses the $FSD$ of premises to order the opinions. This result underscores the importance of the premise in ranking investors' opinions. We also observe an improvement in filtering out opinions with lower MPP when only the key premise is used.

In conclusion, the proposed Fuzzy Strength Degree ($FSD$) proves useful in recommending investor opinions leading to higher MPP. With the introduction of the argument-based opinion analysis concept, both the efficiency of highlighting opinions with higher MPP and of filtering out those with lower MPP have improved.

\subsection{Recommendation Results in SMP Set}
\label{sec:Recommendation Results in SMP set}
Unlike previous research~\cite{zong-etal-2020-measuring,chen2021evaluating}, which focused either on professional reports or social media posts, our study explores both realms. In this section, we employ ExpertLike~\cite{chen2021evaluating} as the benchmark. ExpertLike ranks social media posts by counting the number of sentences in the posts that are classified as professionally written. Their method, which sorts social media posts based on professional-appearing content, has been demonstrated to be profitable.

Building on their approach, we propose an alternative strategy for comparison: 
\begin{itemize}
    \item \textit{ExpertLike + FSD}: Compute the average of the Fine-grained Strength Degrees ($FSD$) of the sentences classified as expert-like.
\end{itemize}
Due to the absence of an additional dataset with Market Price Change Percentage (MPP) labels, we cannot employ supervised models in this section. Moreover, the lack of argument annotations for social media posts precludes testing \textit{AllArg}, \textit{ClaimOnly}, \textit{PremiseOnly}, and \textit{KeyPremise} strategies. Although we attempted to use the pretrained models from \citeauthor{chen2021evaluating}~\cite{chen2021evaluating} for professional report argument analysis, none of the sentences in the SMP set were identified as claims or premises. Thus, we apply the \textit{AllSent} strategy in our experiment, examining whether the proposed $FSD$ also applies to amateur investors' opinions.

\begin{table}[t]
  \centering
    \resizebox{\columnwidth}{!}{
    \begin{tabular}{lrrrr}
    \hline
    \multirow{2}[0]{*}{Strategy} & \multicolumn{2}{c}{Top} & \multicolumn{2}{c}{Last} \\
          & 10th Decile & 9th Decile & 2nd Decile & 1st Decile \\
    \hline
    \textit{ExpertLike + FSD}   & 16.18\% & 12.98\% & \multicolumn{1}{c}{-} & \multicolumn{1}{c}{-} \\
    ExpertLike~\cite{chen2021evaluating} & 17.61\% & 13.09\% & \multicolumn{1}{c}{-} & \multicolumn{1}{c}{-} \\
    \textit{AllSent}   & \textbf{19.41\%} & \textbf{15.79\%} & \textit{12.38\%} & \textit{9.93\%} \\
    \hline
    \end{tabular}%
    }
        \caption{Averaged MPP of the sorted amateur posts based on argument strength under different strategies.}
  \label{tab:Averaged MPP of the sorted amateur posts}%
\end{table}%

Table~\ref{tab:Averaged MPP of the sorted amateur posts} presents the experimental results in the SMP set. Firstly, we find that the top-ranked posts sorted by argument strength yield higher profits than those sorted by other strategies. Secondly, the \textit{ExpertLike + FSD} results indicate that the fusion of previous work and the proposed method successfully ranks the opinions, although the potential profits using this combined strategy are not as high as implementing the two approaches separately. Thirdly, we observe that only 23.16\% of posts in the SMP set contain at least one expert-like sentence as per the ExpertLike approach~\cite{chen2021evaluating}. Consequently, the majority of posts (76.84\%) receive zero scores when using the method proposed in the previous work, which restricts the applicability of their approach in ranking opinions. In contrast, \textit{AllSent} results demonstrate that our proposed method can also be applied to rank social media posts, reinforcing the correlation between proposed $FSD$ and the MPP of amateur opinions.

In conclusion, this section offers experimental evidence validating the effectiveness of the proposed $FSD$ in recommending amateur posts. Furthermore, it highlights the limitations of previous methodologies and introduces an approach capable of ranking all opinions.

\section{Discussion}
\subsection{Risk Analysis}
Risk consideration is a vital aspect of financial decision-making. Proper understanding of associated risks can guide investors in selecting strategies that align with their risk tolerance. In this section, we examine the risk associated with the top-ranked and the last-ranked opinions. We employ the maximum loss (ML) metric, as proposed by prior work~\cite{chen2021evaluating}, to evaluate the risk. ML captures the potential maximum loss if one were to trade following a particular opinion. The ML of bullish and bearish reports are calculated using the following equations:

\begin{equation}
    ML_{\mathit{bullish}} = \min_{t+1 \le i \le t+T}\frac{L_i -O_{t+1}}{O_{t+1}}
\end{equation}
\begin{equation}
    ML_{\mathit{bearish}} = \max_{t+1 \le i \le t+T} \frac{O_{t+1}-H_i}{O_{t+1}}
\end{equation}

Table~\ref{tab:Averaged ML of the sorted professional reports under different strategies} presents the MLs under various strategies. Interestingly, the opinions sorted by the \textit{KeyPremise} not only yield higher profits but also pose lower risks. This pattern is observed in both the top-ranked and last-ranked scenarios, emphasizing the utility of the proposed method in investor opinion recommendation tasks.

However, the experiment results in the SMP set, as shown in Table~\ref{tab:Averaged ML of the sorted amateur posts under different strategies}, indicate that while the proposed approach can discern amateur opinions leading to higher profits, the selected opinions tend to carry higher risks. This introduces a trade-off for investors when using the proposed approach to select social media posts.

\begin{table}[t]
  \centering
   \resizebox{\columnwidth}{!}{
    \begin{tabular}{lrrrr}
    \hline
    \multirow{2}[0]{*}{Strategy} & \multicolumn{2}{c}{Top} & \multicolumn{2}{c}{Last} \\
          & 10th Decile & 9th Decile & 2nd Decile & 1st Decile \\
    \hline
    BERT-Conf~\cite{zong-etal-2020-measuring} & -10.59\% & -10.44\% & -10.38\% & -10.05\% \\
    BERT-Reg~\cite{devlin-etal-2019-bert} & -10.40\% & -10.39\% & -10.58\% & -11.55\% \\
    Mengzi-FinBERT-Reg~\cite{zhang2021mengzi} & -9.70\% & \textbf{-9.95\%} & -11.52\% & -11.42\% \\
    \textit{AllSent}   & -11.30\% & -11.98\% & -9.24\% & -8.80\% \\
    \textit{AllArg}   & -11.48\% & -10.81\% & -10.30\% & -10.39\% \\
    \textit{ClaimOnly}   & -10.70\% & -10.68\% & -10.75\% & -10.50\% \\
    \textit{PremiseOnly}   & -10.53\% & -10.95\% & -12.22\% & -19.95\% \\
    \textit{KeyPremise}   & \textbf{-9.47\%} & \textbf{-9.95\%} & \textit{-15.45\%} & \textit{-22.53\%} \\
    \hline
    \end{tabular}%
    }
    \caption{Averaged ML of the sorted professional reports under different strategies. Because risk is the lower, the better metric, we bold the lowest one for the top-ranked results and use italic to mark the highest one for the last results.}
  \label{tab:Averaged ML of the sorted professional reports under different strategies}%
\end{table}%

\subsection{Behavior of Professional Analysts}
\label{sec:Professional Analysts' Behaviors}
The actions of professional analysts carry significant influence in the financial market, impacting future price movements and other investors' decisions. Conrad et al.~\cite{conrad2006analyst} examined professional analysts' reactions to public information such as news when making recommendations. They found that analysts' recommendations can help predict future price movements. Hirst et al.~\cite{hirst1995investor} found that the strength of arguments in analysts' reports influence investors' judgments. Niehaus et al.~\cite{niehaus2010impact} explored how research coverage by analysts affects the market share of the broker. Inspired by these findings, we aim to examine whether professional analysts and traders have different reactions towards top-ranked and last-ranked opinions.

Unlike previous studies~\cite{zong-etal-2020-measuring,chen2021evaluating} that focused solely on profitability and risk, we propose two novel discussion directions: (1) Do recommended professional reports influence other professional analysts' views in the near future? (2) Do professional traders take actions (buy/sell/no action) in alignment with the recommendations (overweight/underweight/neutral) in the recommended analysts' reports? The analyses in this section aim to elucidate potential future outcomes after following the recommended opinions in trading.

To determine whether recommended professional reports influence other professional analysts' views in the near future, we gather recommendation data from all analysts via the Bloomberg Terminal. We align the reports in the PAR set with the collected data based on the date. For each report in the PAR set, we determine whether any professional analyst changed their view on the same stock in the period from $t+1$ to $t+6$, where $t$ is the report release date. We use $P_{ANA}$, as defined in Equation~\ref{eq:pana}, to compare the reactions of professional analysts to recommended opinions across different strategies.

\begin{equation}
\label{eq:pana}
    P_{ANA} = \frac{N_{RR_i}^f}{N_{RR_i}}
\end{equation}
Here, $RR_i$ represents the recommended reports in the ith decile, $N_{RR_i}$ is the total number of recommended reports in the ith decile, and $N_{RR_i}^f$ represents the number of reports in $RR_i$ that meet the criteria (i.e., at least one professional analyst changes their view on the same stock in the period from $t+1$ to $t+6$).

\begin{table}[t]
  \centering
   \resizebox{\columnwidth}{!}{
    \begin{tabular}{lrrrr}
    \hline
    \multirow{2}[0]{*}{Strategy} & \multicolumn{2}{c}{Top} & \multicolumn{2}{c}{Last} \\
          & 10th Decile & 9th Decile & 2nd Decile & 1st Decile \\
    \hline
    ExpertLike~\cite{chen2021evaluating}    & \textbf{-3.72\%} & \textbf{-6.26\%} & \multicolumn{1}{c}{-} & \multicolumn{1}{c}{-} \\
    \textit{ExpertLike + FSD}   & -4.23\% & -6.38\% & \multicolumn{1}{c}{-} & \multicolumn{1}{c}{-} \\
    \textit{AllSent}   & -10.22\% & -8.33\% & -5.41\% & -7.42\% \\
    \hline
    \end{tabular}%
    }
    \caption{Averaged ML of the sorted amateur posts under different strategies.}
  \label{tab:Averaged ML of the sorted amateur posts under different strategies}%
\end{table}%

\begin{table}[t]
  \centering
  \resizebox{\columnwidth}{!}{
    \begin{tabular}{lrrrr}
    \hline
     & \multicolumn{2}{c}{Top} & \multicolumn{2}{c}{Last} \\
          & \multicolumn{1}{c}{10th Decile} & \multicolumn{1}{c}{9th Decile} & \multicolumn{1}{c}{2nd Decile} & \multicolumn{1}{c}{1st Decile} \\
    \hline
    $P_{ANA}$  & 10.53\% & 10.75\% & 0.00\% & 0.00\% \\
    \hline
    \end{tabular}%
    }
    \caption{Analysis of the professional analysts behaviors after the report released date. The recommendation in this table is based on \textit{KeyPremise} strategy.}
  \label{tab:behavior-view}%
\end{table}%

Table~\ref{tab:behavior-view} provides the statistics of $P_{ANA}$ for reports in different deciles, ranked by the \textit{KeyPremise} strategy. Intriguingly, top-ranked reports are more likely to influence a change in other professional analysts' views compared to last-ranked reports. This finding implies that reports recommended by our proposed approach receive more attention from professional analysts than those at the bottom of the ranking.

\subsection{Professional Traders' Behaviors}
\label{sec:Professional Traders' Behaviors}
Although everyone can express opinions, not all opinions will manifest in market actions. When investors put their money into a specific stock, it shows a strong affirmation of the said stock. Therefore, investors' trading behaviors have long been a focal point of research. 
Chordia et al.~\cite{chordia2001market} examined the relationship between market liquidity and trading activities, discovering increased trading activity preceding macroeconomic information announcements. Furthermore, Chordia and Subrahmanyam~\cite{chordia2001trading} correlated trading activities with expected returns, asserting the importance of trading activity in the cross-section of expected returns. In another study, Wüstenfeld et al.~\cite{wustenfeld2021economic} explored Bitcoin trading activity, suggesting investors in different countries exhibit varied attitudes toward Bitcoin investment. 

Unlike these previous studies, which focus on using market volume-related information to understand investor trading activities, our research targets the trading behaviors of professional traders—those employed in financial institutions. We hypothesize that if professional traders buy (or sell) a stock mentioned in a recommended bullish (or bearish) report, they are likely in agreement with the report's opinions. 

To investigate whether professional traders concur with recommended opinions, we collected publicly available transaction records from the Taiwan Stock Exchange. These records contain details on the number of units that professional institutions buy/sell for a specific stock on a given date. The records are further categorized into three groups: Qualified Foreign Institutional Investors (QFII), Investment Trust (Fund), and Dealer. Using these records, we determined whether these professional traders traded stocks in the same direction as the recommended reports. We calculate the concurring ratio ($CR$) as defined in Equation~\ref{eq:cr}.

\begin{equation}
\label{eq:cr}
    CR = \frac{N_{RR_i}^c}{N_{RR_i}}
\end{equation}
Here, $N_{RR_i}^c$ is the number of times professional traders act in the same direction as the reports in $RR_i$ on the next trading day after the report release. 

Table~\ref{tab:trade_professional} presents the results in the PAR set. We observe that top-ranked opinions garner more favor from professional traders than ranked-last opinions, irrespective of institution type. As QFII's positions represent about 40\% of the total stock market capitalization, their behaviors are considered more influential. Notably, QFII shows a high agreement towards recommended stocks. This suggests our proposed method successfully identifies important premises that align with professional traders' decision-making process.

To sum up, we utilize professional behaviors as an indicator of our proposed method's hit ratio, discussed in Section~\ref{sec:Professional Analysts' Behaviors} and this section. Results indicate that our approach effectively identifies premises aligned with professional perspectives and filters out those that do not pique professional traders' interest.

\begin{table}[t]
  \centering
  \resizebox{\columnwidth}{!}{
    \begin{tabular}{lrrrr}
    \hline
    \multirow{2}[0]{*}{} & \multicolumn{2}{c}{Top} & \multicolumn{2}{c}{Last} \\
          & \multicolumn{1}{c}{10th Decile} & \multicolumn{1}{c}{9th Decile} & \multicolumn{1}{c}{2nd Decile} & \multicolumn{1}{c}{1st Decile} \\
    \hline
    $CR$-QFII & 50.44\% & 50.66\% & 27.81\% & 0.88\% \\
    $CR$-Fund & 34.65\% & 34.43\% & 7.51\% & 15.11\% \\
    $CR$-Dealer & 39.04\% & 41.23\% & 9.27\% & 18.67\% \\
    \hline
    \end{tabular}%
    }
   \caption{Analysis of the professional traders' behaviors after the report released date. Overall means the sum of the units traded by the traders in QFII, fund, and dealer. The recommendation in this table is based on \textit{KeyPremise} strategy.}
  \label{tab:trade_professional}%
\end{table}%

\subsection{Evaluation on Full Recommendation List}
To perform a more granular evaluation in the investor opinion recommendation scenario, Table~\ref{tab:MPP and ML of Top-10 and Top-20 posts} presents the results of average MPP and ML of top-10 and top-20 reports, sorted by each method. \textit{AllSent} performs the best under this setting, while \textit{KeyPermise} ranks second in terms of MPP. From the risk perspective (ML), \textit{AllSent} still outperforms other methods. These findings confirm that \textit{AllSent}, which employs the proposed $FSD$ to rank investor opinions, is the most suitable method for investor opinion recommendation tasks among all methods tested. 

Firstly, despite \textit{AllSent} slightly underperforming \textit{KeyPremise} in MPP in the 10th decile (Top-228 Reports) and the 9th decile (Top-456 reports) as depicted in Table~\ref{tab:Results of sorting professional reports}, the top-10 and top-20 reports recommended by \textit{AllSent} achieve higher MPP and lower ML than those based on other methods. The real-world application scenario likely involves recommending 10 to 20 reports, allowing investors to review these suggestions and make investment decisions based on their experience. 
Secondly, although \textit{AllSent} lags slightly behind \textit{KeyPremise} in filtering out opinions leading to lower MPP as displayed in Table~\ref{tab:Results of sorting professional reports}, the overall ranking results based on nDCG show that \textit{AllSent} is successful in ranking opinions by MPP. Since investor opinion recommendation aims to identify the opinions leading to profitable outcomes, we propose that the \textit{AllSent} strategy is ideally suited for this purpose. 
Thirdly, \textit{AllSent} has proven to be helpful in both professional and amateur investors' opinion recommendations. However, \textit{KeyPremise} requires additional annotation data and is currently unexplored in the context of amateur investor opinion recommendation.

\begin{table}[t]
  \centering
  \resizebox{\columnwidth}{!}{
    \begin{tabular}{lrrrr}
    \hline
          & \multicolumn{1}{c}{Top-10 MPP} & \multicolumn{1}{c}{Top-20 MPP} & \multicolumn{1}{c}{Top-10 ML} & \multicolumn{1}{c}{Top-20 ML} \\
    \hline
    BERT-Conf~\cite{zong-etal-2020-measuring} & 8.84\% & 7.38\% & -11.53\% & -14.11\% \\
    BERT-Reg~\cite{devlin-etal-2019-bert} & 15.91\% & 12.94\% & -12.94\% & -12.60\% \\
    Mengzi-FinBERT-Reg~\cite{zhang2021mengzi} & 13.16\% & 9.90\% & -8.28\% & -9.98\% \\
    \textit{AllSent} & \textbf{18.89\%} & \textbf{17.69\%} & \textbf{-4.05\%} & \textbf{-6.65\%} \\
    \textit{AllArg} & 9.04\% & 12.70\% & -10.73\% & -10.87\% \\
    \textit{ClaimOnly} & 13.28\% & 13.47\% & -11.16\% & -9.61\% \\
    \textit{PremiseOnly} & 11.01\% & 12.76\% & -6.31\% & -6.62\% \\
    \textit{KeyPremise} & 14.75\% & 15.44\% & -12.71\% & -11.72\% \\
    \hline
    \end{tabular}%
    }
    \caption{MPP and ML of Top-10 and Top-20 posts.}
  \label{tab:MPP and ML of Top-10 and Top-20 posts}%
\end{table}%

\section{Conclusion}
In this paper, we have introduced a new approach for estimating strength degrees and proposed methods incorporating argument-based opinion analysis notions. The experimental results have demonstrated the effectiveness of our approach in recommending both professional and amateur opinions. Furthermore, we have provided comprehensive discussions on several important aspects, such as profitability, risk, professionals' behaviors, and evaluation considerations. Our findings can be leveraged in future work to develop recommendation systems for investors or to filter out opinions that may result in low MPP, thereby reducing noise in model inputs.
Future research can investigate cross-lingual applications and extend the concept of strength degree to analyze additional persuasive essays and debates.

\section*{Limitation}
First, our datasets are predominantly one market. This geographical and linguistic limitation might affect the generalizability of our findings to other regions and languages. Further studies should consider incorporating datasets from diverse markets and languages to validate the robustness and applicability of our approach globally.
Second, our analysis of professional analysts' and traders' behaviors is based on publicly available transaction records and recommendation data. However, these records might not capture the full spectrum of professional activities and decision-making processes. More granular data, including intraday trading patterns and proprietary analyst notes, could offer deeper insights into professional behaviors in response to recommended opinions.

\section*{Ethical Considerations} 
The use of systems in financial markets raises ethical concerns, particularly regarding market manipulation and the amplification of biased opinions. It is crucial to implement safeguards and ethical guidelines to ensure that the system promotes fair and unbiased recommendations, preventing any potential misuse.

\bibliography{acl_latex}

\end{document}